%% file: main.tex
\documentclass[Afour,sageh,times]{sagej}
\usepackage{booktabs}
\usepackage{array}
\usepackage{url}
\usepackage{hyperref}
\hypersetup{hidelinks,pdftitle={Measuring climate backlash in Twitter and Reddit archives},pdfauthor={Wentao Xu}}
\graphicspath{{figures/}}
\makeatletter
\renewcommand{\@maketitle}{%
  \vspace*{-12pt}
  {\sagesf\titlesize\bfseries\raggedright\@title\par}
  \vspace{8pt}
  {\sagesf\normalsize\bfseries\centering\@author\par}
  \vspace{3pt}
  {\sagesf\small\itshape\centering
    \href{mailto:myrainbowandsky@gmail.com}{myrainbowandsky@gmail.com}\par}
  \vspace{12pt}
  \noindent\usebox{\absbox}\par
  \vspace{8pt}
  \noindent\@keywords\par
  \vspace{16pt}}
\makeatother
\title{Measuring climate backlash in Twitter and Reddit archives: Lexical definitions, recorded responses and participant turnover}
\author{Wentao Xu}

\begin{document}
\begin{abstract}
Social media archives are often used to study resistance to climate action, but words, response counters and observed participants do not measure the same social process. We examine four supplied Twitter and Reddit archives by processing all registered files without sampling and applying transparent, non-exclusive lexical rules. The study links frame co-occurrence to source-specific temporal and response models, then separates event-period changes among returning authors from participant turnover. Renewable-energy terms accompany cost-related language on Reddit, yet narrower backlash phrases sharply reduce cross-source contrasts and reverse the sign of the Paris Agreement contrast in submissions. Cross-discourse history does not improve eligible primary-context forecasts. Denial/hoax terms are associated with higher recorded Twitter likes, whereas Reddit response associations depend on frame, outcome and author specification. Around the 2019 global climate strike, returning-author expression and participant turnover both contribute to increased protest-language shares. An archive endpoint prevents the corresponding Climate Twitter migration inference. Most crossed-cluster estimates lack released intervals, and joint author/month response covariance estimates fail, restricting formal inference. These results show how operational definitions, platform-specific response fields and observation boundaries shape what can be claimed about climate backlash. The contribution is an archive-based account of these measurement consequences, rather than a measure of individual opposition, persuasion or advocacy-induced backlash.
\end{abstract}
\maketitle
\thispagestyle{plain}

\input{introduction.tex}
\input{conceptual_framework.tex}
\section{Methods}
\input{methods.tex}
\section{Results}
\input{results.tex}

\input{discussion.tex}

\setlength{\bibsep}{0.12em}
\renewcommand{\bibfont}{\small}
\bibliographystyle{SageH}
\bibliography{references}
\label{main:lastpage}
\end{document}

%% file: introduction.tex
\section{Introduction}

An increase in climate-related posts can indicate mobilization, resistance, news attention or a change in the people whose contributions enter an archive. These possibilities matter when social media counts are used to describe a public reaction to climate action. Survey research distinguishes support from perceptions of support, and research on climate beliefs identifies substantial political and social variation \citep{andre2024,hornsey2016}. A platform archive observes a different object: selected contributions and response fields captured under particular collection conditions. The challenge is to explain what those observations establish before treating them as evidence of backlash.

Climate opposition also extends beyond denying warming. Research on political polarization and organized climate communication documents differences in risk perception, policy attitudes and the production of public messages \citep{kahan2012,mccright2011,gustafson2019,farrell2016,supran2017}. Discourses of delay can accept climate change while questioning the feasibility, fairness or timing of action \citep{lamb2020}. A discussion of renewable-energy costs may therefore be relevant to opposition, but a cost term alone can also occur in a proposal, a comparison or a rebuttal. The same ambiguity applies to skepticism and alarmism terms. Counting these words can describe discourse without identifying the writer's position.

Twitter and Reddit offer extensive records of such discussion. Research has examined polarized climate frames, conversation communities and changes around climate events \citep{janghart2015,williamsetal2015,falkenberg2022}. Studies of movement and news actors trace climate-strike framing, while semantic-network research follows the rise of climate-action language \citep{chen2023,suitner2023}. Reddit studies examine deliberation, polarization and long-run changes in terminology \citep{treen2022,fariello2025}. Reviews show that these contributions depend on platform selection and on how a climate public is defined \citep{pearce2019,fownes2018}. These studies motivate comparison across sources, but they do not make lexical adjacency, a like count and an author's changing text interchangeable evidence.

A second line of inquiry examines how data practices produce the objects that researchers then analyse. \citet{kitchin2014} challenges the idea that large datasets eliminate questions of theory and interpretation. \citet{felt2016} connects social media analytics to the collection choices that make records available, and \citet{gray2018} argue for literacy in the infrastructures through which data are produced and used. Archive research further shows that formats and extraction practices affect what can become evidence \citep{maemura2023}. For climate discourse, this perspective directs attention to the practical relation between an asserted social category, its operational rule and the records that survive the collection process.

Critical studies of climate-related data initiatives also caution against taking the vocabulary of climate action at face value. \citet{espinoza2021} analyse how a corporate data-for-climate initiative legitimizes data practices; \citet{riemens2025} studies how technology companies legitimate environmental roles through tech-on-climate discourse. Our archives do not test those corporate strategies. They raise a related measurement question: what follows when heterogeneous climate-action language is counted as a common category, paired with broadly defined backlash terms and interpreted through different platform response fields? Answering requires an empirical account of the consequences of those choices, rather than assuming that more records provide a better measure of public opposition.

We address this question using all registered files in four supplied collections: Climate Twitter, Debunking Twitter, Reddit submissions and Reddit comments. The archive partitions cover portions of 2010--2024, with different observed periods by source. We retain a climate-context denominator independent of the action and backlash labels, analyse the collections separately and report where inferential support is insufficient. Full-corpus processing refers to these supplied archives, not a census of Twitter or Reddit. The contribution is to connect three levels of evidence that are often collapsed: within-record lexical association, source-specific timing and recorded response, and event-period changes in contributors' expression and composition.

The research questions follow that sequence. \textbf{RQ1:} Which action forms dominate the observed environmental discourse, and how do action terms co-occur with backlash-related lexical frames? This establishes the categories and within-source relationships to be compared. \textbf{RQ2:} How do action--backlash relationships vary between Twitter and Reddit over time, and which frames are associated with higher platform-observed responses? This tests whether co-occurrence extends to additional temporal information and response associations without assuming comparable platform outcomes. \textbf{RQ3:} Around major climate, movement or policy events, how do aggregate frame changes relate to expression among returning authors and participant turnover? This asks whether event-period change can be attributed descriptively to the same observed contributors or to a changing archive population. Together, the questions assess how far the available evidence supports a social account of climate backlash.

%% file: conceptual_framework.tex
\section{Platform archives and the measurement of climate backlash}

\subsection{Lexical rules define the measured relationship}

Automated text analysis requires a stated connection between an observable feature and a proposed interpretation \citep{grimmer2013,gentzkow2019}. Here, a frame is a transparent set of applied expressions, not a recovered psychological disposition. Multiple labels can match one record because discussion can connect protest, energy technologies and policy costs. This preserves co-occurrence while allowing quoted opposition, endorsement and rebuttal to enter the same lexical category. It does not resolve those positions.

Climate-communication research gives reasons to maintain that distinction. Analyses of denial and climate-science controversy examine arguments and communicative context rather than treating every related word as an endorsed position \citep{walter2018,jaspal2013,bloomfieldtillery2022}. Our broad backlash-related labels cover several possible topics of resistance, but they also include generic cost and uncertainty words. Explicit phrases provide a narrower alternative definition, not a validated stance benchmark. Comparing the two definitions asks how an empirical contrast changes when the measured vocabulary changes. A difference in magnitude or sign is evidence about operationalization; it cannot by itself determine which definition better represents opposition.

\subsection{Response fields observe different forms of participation}

Research on connective action explains how personalized contributions can participate in collective mobilization without reproducing a single organizational model \citep{segbergbennett2011,bennettsegberg2012}. It does not imply that every visible interaction records mobilization. Twitter likes and Reddit signed scores are different recorded outcomes. Neither field identifies what a recipient saw, why the recipient responded, or whether the response translated into action. \citet{kennedy2015} distinguish making publics known through data mining from supporting publics' own agency. In an archive study, a response count belongs to the former evidential problem; attributing agency requires additional observations.

We therefore estimate response associations within sources. Twitter count models and Reddit transformed-score models answer different statistical questions and retain their different units. A coefficient can identify a conditional association between a word category and an archived response field, subject to the error model. It does not establish popularity across platforms or the persuasive efficacy of an action frame. Temporal models provide another kind of evidence: whether one discourse's past adds information about another's subsequent measured share. That claim concerns prediction, not an observed exposure or an identified causal pathway.

\subsection{Observed contributors delimit event-period change}

An event can coincide with changes in language and changes in who contributes. Social-identity accounts of collective action concern identification, perceived injustice and efficacy \citep{vanderzomeren2008,masson2021}; longitudinal climate-activism research measures psychological and behavioural variables that textual archives do not directly observe \citep{castiglione2022}. The relevant archive question is narrower: does an aggregate increase reflect more action-language use by returning accounts, different posting weights, or the contributions of accounts observed on only one side of the event?

This distinction links participant-level analysis to infrastructure rather than treating author matching as a sufficient solution. A returning author is an account observed in both windows, not a representative participant followed independently of collection. An apparent exit can reflect no captured post, an absent identifier or an archive endpoint. \citet{maemura2023} demonstrates the material consequences of archive construction; our CSV and Parquet collections differ from web-archive formats, but their observation boundaries likewise require examination. Paired text-state analysis and an exact turnover decomposition can account for changes in observed records. They cannot turn incomplete capture into evidence of withdrawal or textual transitions into belief conversion.

These distinctions guide the sequence of analyses. Lexical comparisons establish what the categories capture; temporal and response models examine additional source-specific relationships; event analyses then separate observed expression from observed composition. The inferential exclusions are part of that account. Large record counts do not remove dependence between repeated authors or compensate for a small number of informative months.

%% file: methods.tex
\subsection{Archive construction and observation units}

We read every registered file in four supplied source groups without sampling. The v4 two-pass procedure identified context-matching IDs across all 2010--2024 partitions, retrieved usable versions and selected one version per source before final context filtering. Twitter versions were ranked by valid capture time, ingestion time, source file and version fingerprint. Reddit versions were ranked by source file and fingerprint. Because collections can overlap, we analysed them separately rather than pooling them.

We treated deleted/removed or empty fields as missing, lowercased the remaining text and collapsed repeated whitespace. Submission title and body were separated by a literal field boundary; comments used body text. Primary context required a word-boundary match for \texttt{climate}, \texttt{global warming} or \texttt{greenhouse}. The sensitivity selector also included \texttt{carbon}/\texttt{emissions}. UTC creation time defined periods, and absent months were not zero-filled. Supplementary Table~S1 gives cohort denominators; Sections S1 and S7 document matching, version reconciliation and provenance. The scans cover the supplied archives, not all historical platform activity.

\input{RQ1_METHODS.tex}

\subsection{Common-month conditional contrasts}

We paired Twitter and Reddit source-months containing at least 20 action and 20 non-action records per source. We then compared equation~\ref{eq:rq1-conditional-difference} under broad labels and explicit phrases. Equally weighted mean Twitter-minus-Reddit contrasts used calendar-distance Bartlett HAC(6), with four-test BH families per definition. Consecutive-run and broader-context analyses assessed sensitivity. Archive selection and community differences preclude causal platform comparisons.

\subsection{Source-specific response regression}

We modelled original-post Twitter likes with negative-binomial NB2 fits that met the numerical acceptance criteria. Rejected Poisson fits were not interpreted. Reddit signed scores used $\operatorname{asinh}(y)=\log(y+\sqrt{y^2+1})$ ordinary least squares. The transform preserves sign and compresses extreme values. Its coefficients are conditional differences on a transformed scale, not raw net-score differences or support proportions.

Models jointly included 16 lexical indicators, month indicators, source-file/subreddit controls, log text length and available Twitter capture age. Author-clustered intervals were primary, with separate month-clustered intervals assessing sensitivity. Each variance method had its own 61-test BH family; three Debunking Twitter coefficients were withheld. All four two-way author/month covariance matrices failed positive-semidefiniteness. The separate sensitivities therefore do not provide valid joint author/time inference. Figure~\ref{fig:rq2-response} displays both intervals and their support classifications; Supplementary Table S6 reports coefficients, $p$ and $q$. Repeat-author fixed-effect models with transformed outcomes formed a separate family. Twitter count ratios and Reddit coefficients have different units.

\subsection{Conditional Granger tests and chronological prediction}

For each source/definition, we used the longest consecutive monthly run with 20 action and 20 non-action records per month and at least 49 months. First-differenced Jeffreys logit shares, $z_t=\Delta\log[(n_t+0.5)/(N_t-n_t+0.5)]$, and log-volume changes required ADF $p<0.05$ and KPSS $p>0.05$. These diagnostic criteria do not prove stationarity. The conditional Granger model was
\begin{equation}
z^B_t=c+\sum_{j=1}^{L}\alpha_j z^B_{t-j}
+\sum_{j=1}^{L}\beta_j z^A_{t-j}+\gamma v_{t-1}+s_t+\epsilon_t,
\end{equation}
where $v_{t-1}=\Delta\log N_{t-1}$ and $s_t$ contains annual sine/cosine terms. The reverse model exchanges $A,B$; the joint null is $\beta_1=\cdots=\beta_L=0$ \citep{granger1969}. The exploratory analysis treats $L=3$ as principal and $L=1,6$ as sensitivities. Approximate Wald $F$ tests use Bartlett HAC(6) \citep{neweywest1987}; Breusch--Godfrey(6) $p<0.05$ flags residual dependence. For each context, BH/BY cover 16 principal or 32 sensitivity tests, with withheld tests entering as $p=1$. Supplementary Section S3 gives fit-support and diagnostic rules.

Expanding-window $L=3$ forecasts used 36 initial training rows and at least 24 evaluation months. Eligibility diagnostics used initial training observations, although run selection remained retrospective. The baseline contained target history, seasonality and volume; the augmented model added cross-history. All predictors preceded outcomes. RMSE reduction was $100[1-\mathrm{RMSE}_{\mathrm{aug}}/\mathrm{RMSE}_{\mathrm{base}}]$. One-sided Clark--West tests used approximate normal HAC(3) inference \citep{clarkwest2007} and separate 16-test BH/BY families. Paired circular blocks gave 95\% intervals conditional on fixed forecasts, excluding refitting uncertainty. Supplementary Section S3 specifies block size, draws, seed and diagnostic windows; metadata records SciPy 1.15.3/statsmodels 0.14.4. These forecasts assess prediction rather than an exposure effect.

\subsection{Segmented event-time and paired-author analysis}

Eight UTC anchors covered Paris adoption/US withdrawal, the IPCC 1.5-degree report, the 2019 strike, the WHO pandemic declaration (attention comparator), COP26 closing, Inflation Reduction Act signing and COP28 opening. Excluding each anchor day, descriptions used 90-day windows. Segmented conditional differences used 26 seven-day bins per side, requiring 20 action and 20 non-action records per bin. Local trend, post-event level/slope and annual sine/cosine terms were fitted with $n_A n_{\neg A}/n$ weights and HAC(4). HAC(8), equal weights and shorter windows were sensitivities. Level departures are not isolated event effects \citep{bernal2017}.

Returning identifiable authors had records in both 90-day windows. Exact two-sided McNemar tests compared gains/losses in 16 frame-presence indicators. Conservative paired intervals differenced two Bonferroni-adjusted exact binomial marginal intervals. We analysed all-returning and two-records-per-side cohorts separately; the activity restriction does not equalize posting opportunities. A distinct modal-state analysis used neither, action-only, backlash-only, mixed and tied states. Bowker/Stuart--Maxwell tests assessed symmetry/marginal homogeneity where supported. These analyses measure text transitions, not belief change.

The strike protest-share change was decomposed into returning-author expression, record weights, entrants, exits and missing authors. For returner $i$, $p_{it}$ is the author's frame share and $w_{it}$ the author's fraction of all records. The accounting identity
\begin{equation}
\Delta(w_i p_i)=\bar w_i\,\Delta p_i+\bar p_i\,\Delta w_i
\end{equation}
separates returning-author components without causal interpretation. Event-level and activity-specific author tests used separate BH/BY families. Placebos at $\pm365,\pm730$ days are diagnostic, not randomized controls; positive daily counts do not certify complete capture.

%% file: RQ1_METHODS.tex
\subsection{Lexical estimands and clustered uncertainty}\label{sec:rq1-v4-methods}

Nine action labels captured protest/strike, activism, renewable energy, electric vehicles, adaptation, mitigation/decarbonization, net zero, energy transition and sustainability. Seven broad backlash-related labels captured denial/hoax, skepticism, alarmism, anti-green, cost-related, control/freedom and general backlash terms. Generic cost and uncertainty words qualify without negation or endorsement rules. We therefore describe the unchanged \texttt{cost\_objection} field as cost-related terms. Any-action $A$ includes ``climate action'' in addition to the nine-label union; broad $B$ is the seven-label union. Explicit $B$ restricts matching to climate hoax/scam/alarmist, anti-green and net-zero sceptic/skeptic phrases and their applied variants. Supplementary Section S1 supplies all applied expressions and preprocessing rules.

The non-exclusive labels measure lexical presence across a record, including quotations and rebuttals, rather than stance. Full-corpus descriptions retain missing authors, while author-clustered inference excludes them. We analyse each source separately.

Each source has 16 frame prevalences, 36 paired action-frame differences and 63 action/backlash conditional differences. For action $A$ and backlash $B$, the latter is
\begin{equation}
 D_{AB}=\Pr(B=1\mid A=1)-\Pr(B=1\mid A=0).
 \label{eq:rq1-conditional-difference}
\end{equation}
The groups and denominator are source-specific. Action-pair differences preserve within-record dependence by differencing binary labels on each record. Neither estimand identifies a causal effect or endorsement.

We calculated crossed author/month variances from linearized influences, subtracting the intersection component and applying finite-cluster corrections \citep{cameron2011}. Two-sided tests and pointwise 95\% intervals used a $t$ reference with $\min(G_{\mathrm{author}},G_{\mathrm{month}})-1$ degrees of freedom. Undefined/nonpositive variances were withheld. Author-only, month-only and calendar-month influence HAC(3,6,12) were fixed sensitivities; Supplementary Section S2 gives the formulas.

The exploratory support screen required 30 authors and 12 months per relevant arm, with an effective squared-score cluster count $\geq10$. No cluster could exceed 25\% of squared influence, and leave-one-cluster estimates had to remain defined. This post-inspection screen does not guarantee coverage. Prevalence has no zero-null test. For each variance method, Benjamini--Hochberg (BH) and Benjamini--Yekutieli (BY) adjust 144 action-difference or 252 conditional tests \citep{benjamini1995,benjamini2001}. Withheld tests enter as $p=1$, without displayed $p/q$. All estimates and reasons for withholding are retained.

%% file: results.tex
\begin{figure*}[t]
\centering
\includegraphics[width=0.85\textwidth]{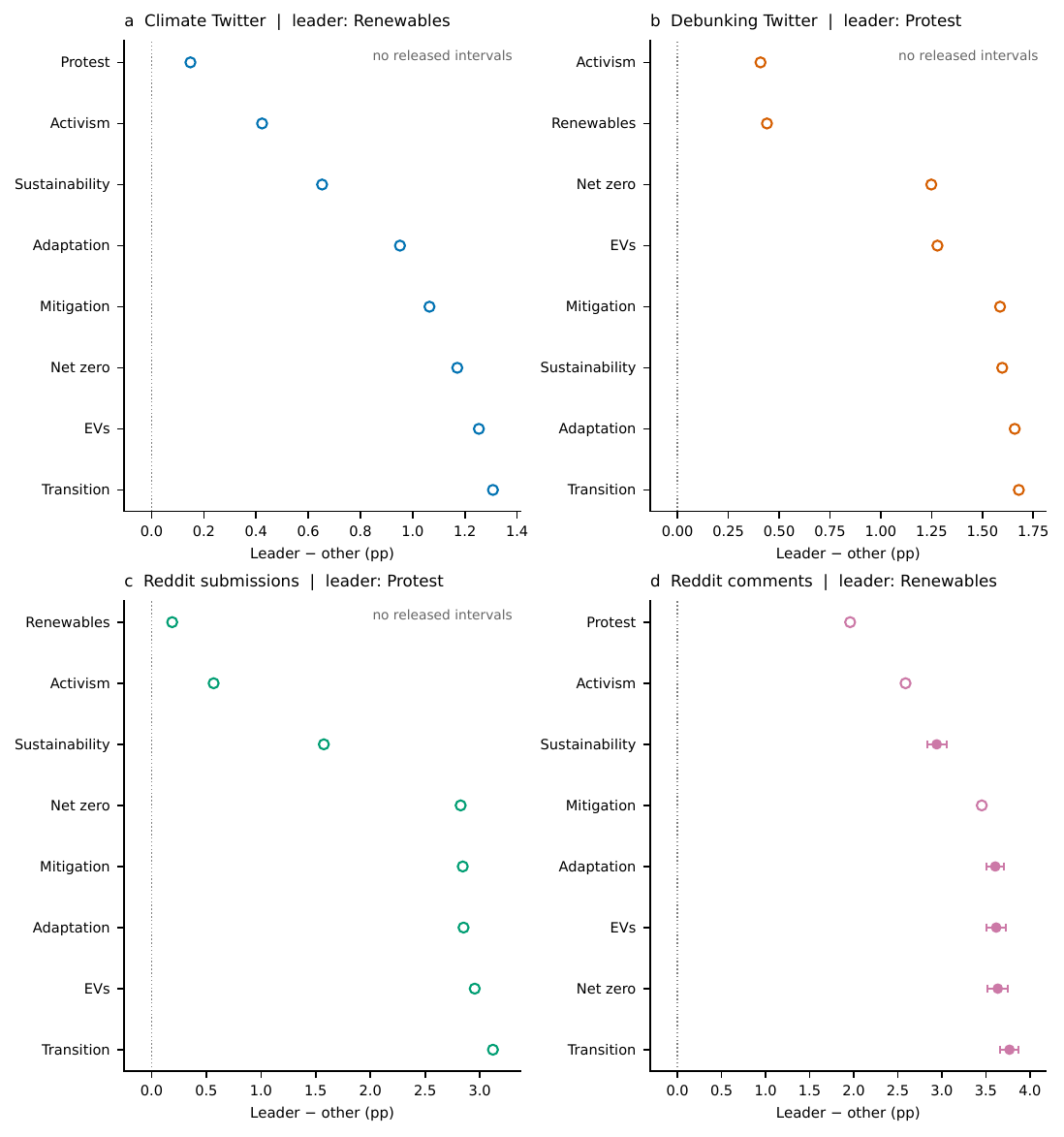}
\caption{Differences between each source's observed leading action frame and its other action frames. Open points lack an interval because crossed author/month inference failed a support or variance criterion. Bars are unadjusted pointwise 95\% intervals; data-selected leaders have no rank-selection adjustment. Horizontal scales differ.}
\label{fig:rq1-cluster-action-dominance}
\end{figure*}

\input{RQ1_RESULTS.tex}
\input{RQ1_conditional_table.tex}

\begin{figure*}[t]
\centering
\includegraphics[width=0.98\textwidth]{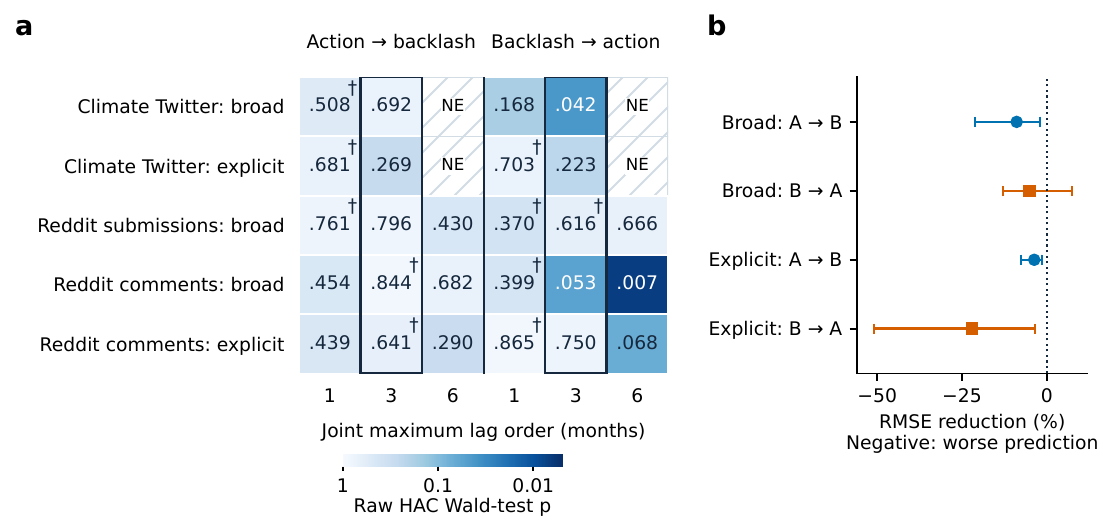}
\caption{Additional prediction information from the other discourse's history. a, Raw approximate HAC(6) Wald $p$ for cross-lags 1 through $L$, controlling own history, seasonality and lagged volume. Outlines mark the principal $L=3$ model; $\dagger$ flags residual-dependence diagnostics. No test passed BH in either the 16-test principal or 32-test sensitivity family. Runs contain 69 Climate Twitter, 169 submission and 98 comment months. Hatched NE indicates a test was not estimated, and does not imply no relationship. Debunking Twitter lacked a sufficiently long run; submission explicit phrases failed stationarity diagnostics. b, Climate Twitter root-mean-square error (RMSE) reduction in 29 evaluation months, with conditional 95\% paired-block intervals; negative values mean worse prediction. Reddit forecasts failed training-only diagnostics. A denotes the any-action selector and B the indicated backlash definition. Expanded-context sensitivities appear in text and source data. Arrows indicate prediction direction, not causation.}
\label{fig:rq2-time}
\end{figure*}

\begin{figure*}[t]
\centering
\includegraphics[width=0.98\textwidth]{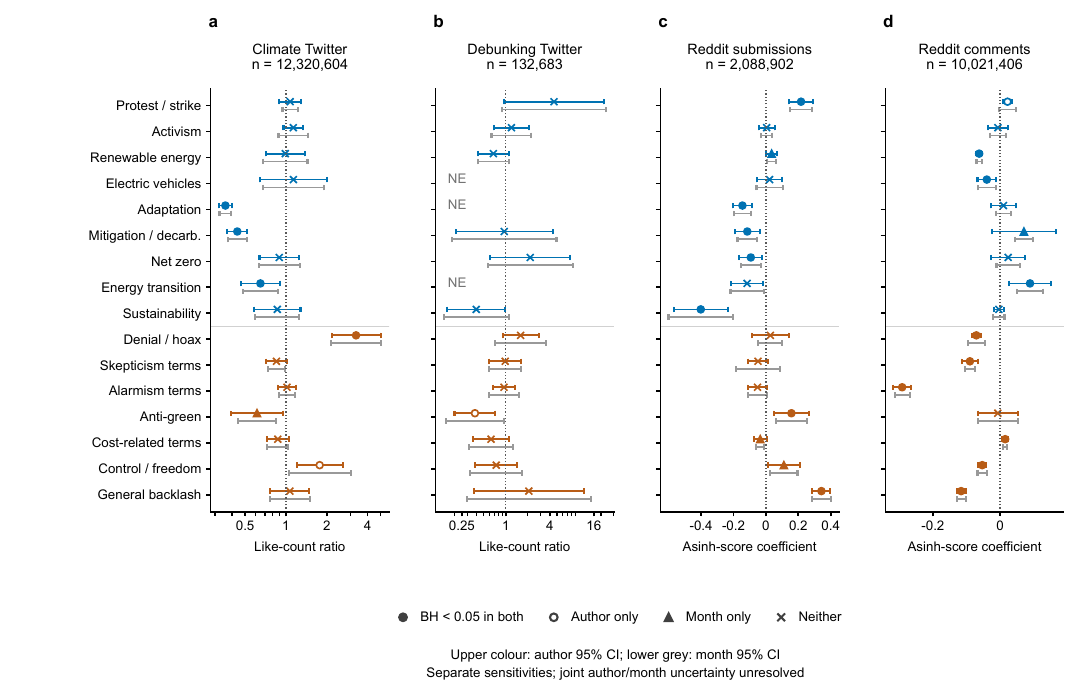}
\caption{Response associations under separate author- and month-clustering sensitivities. a,b, Original-Twitter-post like-count ratios from NB2 fits ($n=12{,}320{,}604$ and $132{,}683$). c,d, OLS coefficients on Reddit $\operatorname{asinh}$ signed scores ($n=2{,}088{,}902$ and $10{,}021{,}406$). Upper coloured intervals use author clustering; lower grey intervals use month clustering. Both are pointwise 95\% intervals. Filled circles pass BH $q<0.05$ in both separate 61-test families, open circles in the author family only, triangles in the month family only, and crosses in neither. NE marks three withheld Debunking Twitter coefficients. This classification does not establish jointly valid author/month uncertainty: all four joint covariance matrices failed. Blue denotes action terms and orange backlash-related terms. Outcome units differ; lexical labels establish neither stance nor causation.}
\label{fig:rq2-response}
\end{figure*}

\subsection{Explicit phrases yield smaller source contrasts}

The mean relationship between action and backlash differed across supported common months and depended on the backlash definition (Supplementary Table~S3). With broad labels, Climate Twitter's monthly conditional difference was 13.53 percentage points lower than Reddit submissions' in 111 common months (95\% HAC interval $-15.29$ to $-11.77$; $p=6.22\times10^{-29}$; BH $q=1.24\times10^{-28}$). It was 14.57 points lower than Reddit comments' in 101 months (95\% interval $-15.65$ to $-13.48$; $p=2.81\times10^{-47}$; $q=1.12\times10^{-46}$). Debunking Twitter was lower than submissions by 8.96 points in 76 months ($q=5.05\times10^{-7}$) and lower than comments by 9.93 points in 64 months ($q=9.10\times10^{-6}$). These estimates compare conditional lexical associations, not total backlash volume or causal platform effects.

Explicit backlash phrases produced much smaller contrasts. Climate Twitter versus submissions was +0.044 percentage points (95\% HAC interval $-0.011$ to +0.099; $p=0.116$, BH $q=0.134$), whereas Climate Twitter versus comments was $-0.069$ points (95\% interval $-0.092$ to $-0.046$; $q=1.57\times10^{-7}$). This definition sensitivity prevents interpreting broad markers as a validated measure of opposition. The consecutive-run sensitivity withheld the Debunking Twitter contrasts because support was insufficient.

\subsection{Cross-discourse history fails to improve eligible primary forecasts}

Contemporaneous associations did not establish additional prediction information. Primary-context Granger tests used 69 consecutive Climate Twitter months, 169 submission months and 98 comment months (Fig.~\ref{fig:rq2-time}a; Supplementary Table~S4). Ten of 16 principal three-month tests were estimable; none passed BH/BY. Climate Twitter's broad backlash-to-action test gave $p=0.0418$, BH $q=0.426$; the reverse gave $p=0.692$. Corresponding comment tests gave $p=0.0533$ and $p=0.844$. Three principal models had residual warnings. Debunking Twitter's 33-month run was too short; submission explicit phrases failed stationarity. These diagnostic exclusions are not null results.

Only Climate Twitter passed primary-context forecast diagnostics (29 evaluation months; Fig.~\ref{fig:rq2-time}b). All four cross-history additions worsened realized RMSE. Action-to-broad-backlash prediction changed by $-8.89\%$ (conditional 95\% interval $-21.08$ to $-1.98$; Clark--West $p=0.940$), whereas the reverse changed by $-5.13\%$ with an interval spanning zero ($-12.99$ to +7.34; $p=0.365$). The explicit-phrase directions changed by $-3.70\%$ and $-22.08\%$. All four BH $q=1.000$; Supplementary Table S5 gives the full intervals and tests. Reddit forecasts were ineligible because they failed training-only diagnostics; they did not undergo a forecast-gain test.

Expanded context produced a six-month broad comment backlash-to-action association ($p=0.000368$, BH $q=0.0118$, BY $q=0.0478$; 91 model months; residual $p=0.979$), but no three-month test passed BH. Of six eligible expanded-context forecasts, only that source/direction improved realized RMSE (+0.38\%, conditional interval $-3.54$ to +4.14; Clark--West $p=0.130$, BH $q=1.000$). The longer-lag result thus depends on the context selector. The three-month forecast check neither directly validates nor refutes the six-month model.

\subsection{Denial and alarmism terms show contrasting response associations}

Response models linked different lexical frames to different archived outcomes (Fig.~\ref{fig:rq2-response}; Supplementary Table~S6). In 12,320,604 eligible original Climate Twitter posts, denial/hoax language was associated with 3.29 times the expected likes (author-clustered 95\% interval 2.17 to 4.99; $p=2.04\times10^{-8}$; BH $q=1.25\times10^{-7}$), whereas adaptation had a like-count ratio of 0.358 (95\% interval 0.319 to 0.400; $p=1.09\times10^{-70}$; $q=3.33\times10^{-69}$). In 2,088,902 submissions, protest/strike had a +0.217 coefficient on transformed signed score (95\% interval 0.140 to 0.293; $p=2.77\times10^{-8}$; $q=1.54\times10^{-7}$); in 10,021,406 comments with valid scores, alarmism had a $-0.292$ coefficient (95\% interval $-0.318$ to $-0.266$; $p=1.62\times10^{-106}$; $q=9.91\times10^{-105}$). These selected associations also passed the separate month-clustered BH family, but the outcomes cannot be directly ranked across platforms.

Clustering choice changed which coefficients passed multiplicity correction. Of 61 estimable coefficients, 20 passed BH in both separate sensitivities, three passed only under author clustering, five only under month clustering and 33 under neither. Comment protest/strike changed from author-clustered $q=0.00101$ to month-clustered $q=0.1387$; Climate Twitter control/freedom changed from 0.0140 to 0.0745, and Debunking Twitter anti-green from 0.0116 to 0.0788. These three associations had clustering-sensitive support. The separate sensitivity checks do not repair the failed joint covariance matrices or establish stable joint inference.

The within-author sensitivity further limited interpretation. Comment alarmism remained negatively associated with score ($\beta=-0.0621$, 95\% interval $-0.0733$ to $-0.0508$; BH $q=5.39\times10^{-26}$). Denial/hoax changed from a negative primary coefficient to a positive within-author coefficient ($\beta=+0.0210$, 95\% interval +0.0113 to +0.0307; $p=2.26\times10^{-5}$; $q=8.89\times10^{-5}$). Sample restriction and author controls changed together, so the reversal does not identify an individual mechanism. Recorded response and directional prediction address different outcomes; neither estimates advocacy's effect on backlash.

\begin{figure*}[t]
\centering
\includegraphics[width=0.90\textwidth]{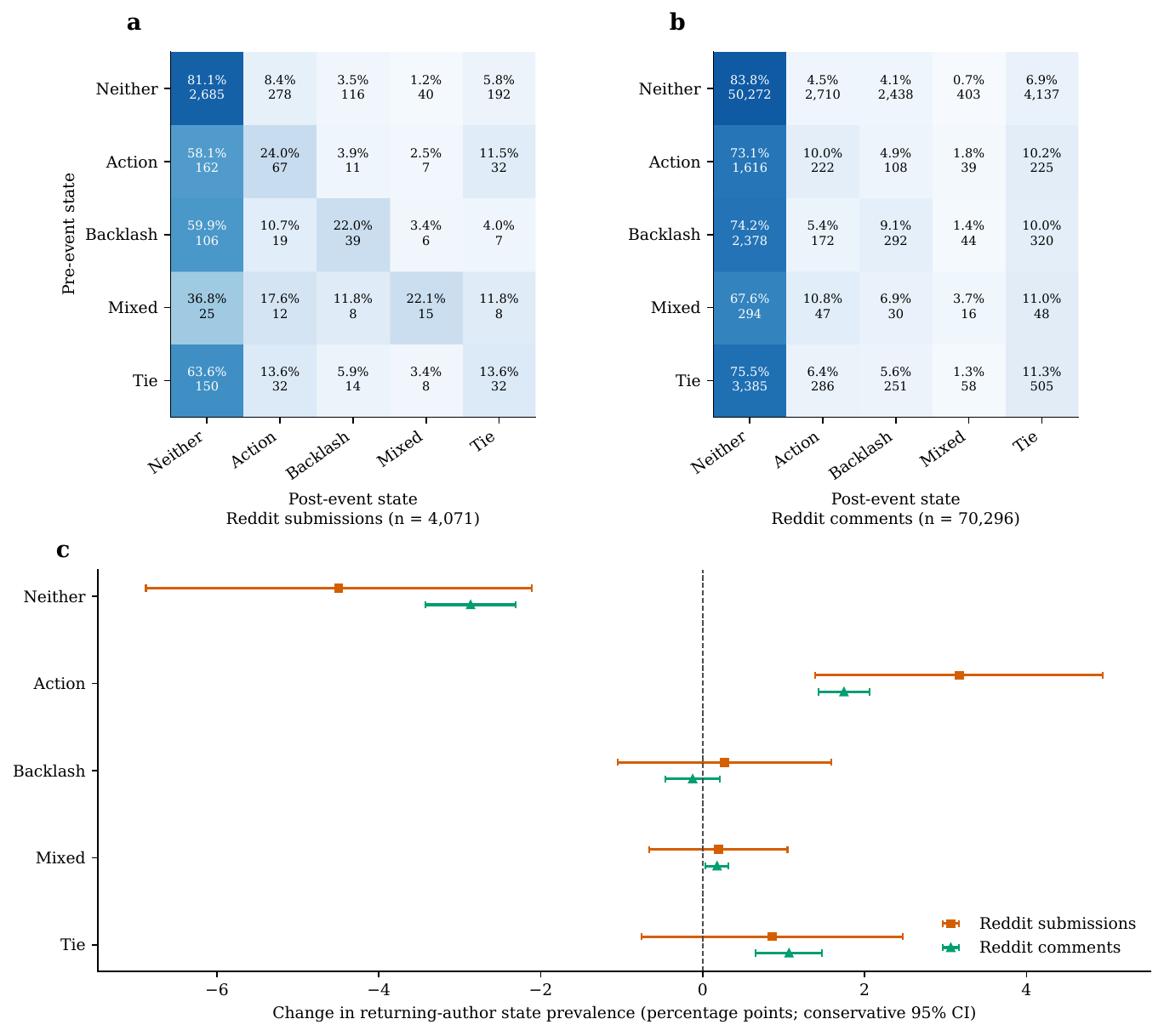}
\caption{Returning-author modal-text transitions around the 20 September 2019 global strike. a,b, Transition counts and row proportions; c, Paired post-minus-pre state prevalence with conservative 95\% exact marginal Bonferroni intervals. Windows are 90 days per side, excluding the event day. Backlash uses the broad seven-label union. A record is Neither when both the any-action selector and backlash union are absent, Action when action alone is present, Backlash when backlash alone is present, and Mixed when both are present. Author state is the uniquely most frequent record state; Tie denotes a tied maximum. Neither is not a neutral stance. Submission returners are 4,071 of 29,516 identifiable pre-event authors (13.79\%); comment returners are 70,296 of 221,136 (31.79\%). These selected cohorts do not represent all event-period participants or belief changes.}
\label{fig:rq3-migration}
\end{figure*}

\subsection{Lexical definitions reverse the Paris Agreement contrast}

Event-local conditional changes depended on the measured words. Paris Agreement adoption had a +9.29-percentage-point Reddit-submission level contrast under broad labels but $-0.445$ points under explicit phrases; both passed BY with their full intervals and tests in Table~\ref{tab:rq3-events} (complete eligible family: Supplementary Table~S7). Equal-week weighting removed broad-label BH support but retained the explicit result. COP26 broad contrasts had opposite signs in submissions and Climate Twitter, neither passing BY; the synchronized-source contrast also failed BH ($q=0.0617$). No general post-event backlash increase follows from these comparisons.

\begin{table*}[t]
\centering
\caption{Selected event-local action--backlash conditional level contrasts. Effects are percentage points from 26-week-per-side segmented models; intervals are pointwise 95\% HAC(4). The BY family includes all 22 eligible primary conditional level tests. Each estimate concerns its own archive and lexical definition, not an event treatment effect.}
\label{tab:rq3-events}
\small\setlength{\tabcolsep}{4pt}
\begin{tabular}{lllrll}
\toprule
Event & Source & Backlash rule & Level [95\% CI], pp & $p$ & BY $q$ \\
\midrule
Paris Agreement & Reddit submissions & broad & +9.29 [3.89, 14.68] & 0.00116 & 0.0470 \\
Paris Agreement & Reddit submissions & explicit phrases & $-0.445$ [$-0.613$, $-0.276$] & $3.17\times10^{-6}$ & $2.57\times10^{-4}$ \\
COP26 & Reddit submissions & broad & +10.99 [3.86, 18.12] & 0.00328 & 0.0851 \\
COP26 & Climate Twitter & broad & $-7.83$ [$-13.06$, $-2.60$] & 0.00419 & 0.0851 \\
\bottomrule
\end{tabular}
\end{table*}

\subsection{Returning authors and turnover account for increased protest shares}

Event-local aggregate change also depended on who contributed records. Around the 20 September 2019 global strike, any-action language rose from 13.09\% to 19.09\% in submissions and from 8.06\% to 12.03\% in comments across 90-day windows. Protest/strike shares increased by 4.299 and 2.405 percentage points. Among 4,071 returning submission authors, the fraction with at least one protest/strike record rose by 7.664 points (conservative 95\% interval 5.538 to 9.776; exact McNemar $p=2.44\times10^{-27}$; BH $q=1.38\times10^{-26}$). Among 70,296 returning comment authors, it rose by 5.729 points (95\% interval 5.311 to 6.145; $p<10^{-300}$; BH $q<10^{-300}$). The two-records-per-side restriction gave increases of 13.493 and 9.015 points. Activity increases opportunities to observe a label, so these are not dose-response tests. All 16 paired-frame results appear in Supplementary Table~S8 and the companion source data.

The five-state analysis detected changes in returning authors' most frequent text state (Fig.~\ref{fig:rq3-migration}). Action-only prevalence increased by 3.169 points in submissions and 1.745 in comments; Supplementary Table~S9 reports the all-returning paired intervals/tests. The submission increase fell to 1.765 points under the two-record restriction, with an interval crossing zero and BH $q=0.0837$; the comment result retained support. Bowker tests gave $\chi^2(10)=45.72$ ($p=1.61\times10^{-6}$) and $404.22$ ($p=1.19\times10^{-80}$), respectively. Changes in an author's modal text state differ from ever mentioning protest. Neither measures persuasion.

Participant turnover also contributed to the aggregate protest/strike increase (Supplementary Table~S10). In submissions, returning-author expression-rate change contributed +2.417 percentage points, record-weight change +0.326, entrants +2.387, exits $-1.308$ and missing-author records +0.477. The contributions summed to +4.299 points. The corresponding comment contributions were +1.076, +0.033, +2.032, $-0.805$ and +0.069, summing to +2.405 points. This exact accounting does not estimate belief change.

Archive coverage constrained strike comparisons. Climate Twitter had 3,171,335 records before the strike but only 5,354 after it; 0.21\% of pre-event authors reappeared. Its largest input archive ended on 19 August 2019, 32 days before the strike. We withheld Climate Twitter strike-migration inference; Debunking Twitter lacked daily support. This exclusion was post hoc. None of 30 fixed-calendar placebo models yielded an adjusted level rejection, but non-rejection cannot identify an event effect or exclude other news.

%% file: RQ1_RESULTS.tex
\subsection{Renewable and protest terms lead the observed action rankings}\label{sec:rq1-v4-results}

After primary climate-context selection, the archives contained 51,508,053 Climate Twitter records, 1,174,002 Debunking Twitter records, 2,259,766 Reddit submissions and 10,279,645 Reddit comments. They spanned 155, 77, 169 and 152 observed months, respectively. All eligible records entered descriptions. Author identifiers were unavailable for 170,864 submissions (7.56\%) and 258,228 comments (2.51\%). These records remained in descriptions but not author-clustered inference. Cost-related terms occurred in 4.16\% of known-author submissions but 1.60\% of missing-author submissions, so the inferential cohort cannot represent the full corpus. The Twitter archives can overlap.

Among records with known authors, renewable-energy language had the highest observed action-frame prevalence in Climate Twitter (1.335\%) and Reddit comments (3.797\%). Protest/strike led in Debunking Twitter (1.689\%) and Reddit submissions (3.204\%). The observed lead over the second-ranked frame ranged from 0.15 percentage points in Climate Twitter to 1.96 points in Reddit comments (Fig.~\ref{fig:rq1-cluster-action-dominance}). None of these four data-selected leader/runner-up contrasts passed the cluster-support screen. The rankings describe the observed records. They do not establish that a leading frame would remain first under the stated repeated-discourse model.

\subsection{Renewable-energy terms accompany cost terms on Reddit}

Among frequent pairs with interpretable inference, technological action terms had the strongest association with cost-related language. In known-author Reddit submissions, cost-related labels were 17.47 percentage points more prevalent in renewable-energy-labelled records than in other records (pointwise 95\% crossed-cluster interval 16.30--18.63; $p=1.88\times10^{-68}$; Benjamini--Yekutieli $q=7.25\times10^{-66}$). The corresponding comment difference was 14.61 points (95\% interval 14.21--15.02; $p=3.00\times10^{-118}$; $q=4.63\times10^{-115}$; Table~\ref{tab:rq1-conditional}). These within-source lexical differences do not show that renewable energy elicited opposition or that the platform caused the difference.

Support differed across archives and variance specifications. Climate Twitter's mitigation/decarbonization and cost-related contrast was 1.32 percentage points under crossed clustering ($q_{\mathrm{BY}}=0.0448$). The specified 3-, 6- and 12-month HAC analyses gave $q_{\mathrm{BY}}=0.919$, 1.00 and 1.00; the 12-month interval spanned $-0.15$ to 2.80 points. Its more frequent renewable-energy and cost-related joint match ($n=18{,}458$) had no released interval because month-level influence was concentrated. The Debunking Twitter pair in Table~\ref{tab:rq1-conditional} had only eight joint records and remains descriptive.

Of 460 crossed-cluster quantities, 67 had released intervals (nine prevalences and 58 contrasts); 393 remained descriptive. Supplementary Table~S2 summarizes releases; Section S2 and source data document withholding rules and reasons. Given this restricted support, we next examined whether lexical definitions changed contemporaneous source contrasts, before testing predictive timing.

%% file: RQ1_conditional_table.tex
\begin{table*}[t]
\centering
\caption{Selected action/backlash conditional prevalence differences, in percentage points. Pairs were chosen by joint frequency, not \(p\). Pointwise 95\% crossed author/month intervals are unadjusted; \(q_{\mathrm{BY}}\) covers 252 tests. The Climate Twitter result does not retain BY \(q<0.05\) under 3-, 6- or 12-month HAC checks. The Debunking Twitter pair has only eight joint records, so its test is not interpreted here. Labels indicate words, not endorsement.}
\label{tab:rq1-conditional}
\small\setlength{\tabcolsep}{3pt}
\begin{tabular}{>{\raggedright\arraybackslash}p{28mm}>{\raggedright\arraybackslash}p{40mm}rrrr}
\toprule
Source & Action / backlash & Joint \(n\) & \shortstack{Difference [95\% CI]\\percentage points} & \(p\) & \(q_{\mathrm{BY}}\) \\
\midrule
Climate Twitter & mitigation / cost-related & 4,151 & +1.32 [+0.55, +2.09] & \(9.01\times10^{-4}\) & 0.0448 \\
Debunking Twitter & protest/strike / cost-related & 8 & -0.61 (descriptive) & \textemdash & \textemdash \\
Reddit submissions & renewable energy / cost-related & 13,294 & +17.47 [+16.30, +18.63] & \(1.88\times10^{-68}\) & \(7.25\times10^{-66}\) \\
Reddit comments & renewable energy / cost-related & 71,307 & +14.61 [+14.21, +15.02] & \(3.00\times10^{-118}\) & \(4.63\times10^{-115}\) \\
\bottomrule
\end{tabular}
\end{table*}

%% file: discussion.tex
\section{Discussion}

The study shows that the interpretation of climate backlash changes with the vocabulary counted, the response field modelled and the contributors observed. Renewable-energy language accompanies cost-related terms on Reddit, but narrower backlash phrases yield much smaller source contrasts and reverse the sign of the Paris Agreement contrast in submissions. Additional cross-discourse history fails to improve eligible primary-context forecasts. Around the global climate strike, protest-language increases include changes among returning authors and contributions from participants observed on only one side. These findings support an account of lexical association and archive composition; they do not establish that advocacy produces opposition.

\subsection{Changing the definition changes the comparison}

The strongest supported co-occurrence differences connect renewable-energy terms with cost-related language. This is substantively relevant because debates about transition policies include distributive and feasibility questions, and climate-delay discourse can operate without outright denial \citep{lamb2020}. Yet our records do not establish whether the matched cost language argues against action, defends it or compares alternatives. Calling the category cost-related terms preserves that uncertainty in the reported result rather than assigning opposition through the field name.

The common-month and Paris Agreement comparisons show a further consequence. A smaller explicit-phrase contrast does not amount to successful validation of the broad labels. The two definitions count different vocabularies. Their difference can alter the social account attached to the result, including whether an event-period relationship appears to increase or decrease. This makes operationalization part of the finding, consistent with criticism of theory-free accounts of data \citep{kitchin2014} and attention to the infrastructures of data use \citep{gray2018}. The empirical contribution is the demonstrated consequence of specific rules within these archives, not a claim that all computational measures are equally uncertain.

These findings complement analyses of corporate climate discourse. \citet{espinoza2021} and \citet{riemens2025} explain how data and technology organizations legitimate environmental roles. Their evidence concerns institutional strategies. Our archive estimates neither identify those actors' strategies nor measure platform companies' environmental claims. They show why a shared climate-action label does not authorize a shared interpretation of intention. Treating renewable energy, protest and corporate climate language as distinct communicative objects remains necessary even when the same expressions occur across them.

\subsection{Recorded response does not identify persuasive success}

Denial/hoax terms have a positive like-count association in the eligible Climate Twitter model, while comment alarmism has a negative transformed-score association. These contrasts describe different response fields and different contributor populations. They cannot be ordered on a common scale of public approval. The within-author comment analysis is especially instructive: the denial/hoax association changes sign when the sample and author controls change together. An explanation based only on the word category would miss the importance of who contributes and which observations enter the model.

Studies of climate polarization, bots and online communities provide reasons to investigate these source differences \citep{falkenberg2022,marlow2021,williamsetal2015}. Our models do not identify bot activity, algorithmic recommendation or network exposure. Subreddit and source-file controls address observed grouping, not those unobserved processes. The distinction drawn by \citet{kennedy2015} between representing a public and enabling its agency also limits what can be inferred from response counters: recorded attention cannot stand in for a recipient's participation in collective action.

The chronological analyses restrict a stronger transmission account. No principal-context conditional Granger test passes the stated multiplicity adjustments, and cross-history additions worsen all eligible primary forecasts. This does not establish that advocacy and backlash are unrelated. The expanded-context six-month comment association prevents that generalization, and diagnostic exclusions leave some relationships untested. The results establish a narrower point: contemporaneous co-occurrence does not provide a dependable basis for claiming additional predictive information under the eligible principal specifications. They also do not identify a causal effect. Studies of fear appeals, motivated reasoning, norms, consensus and inoculation use evidence about responses to messages or interventions beyond lexical presence \citep{oneill2009,feinberg2011,hart2012,sparkman2017,vanderlinden2015,spampatti2024}. A causal tree, forest or language model would not supply the missing treatment definition and identification assumptions.

\subsection{Participant turnover contributes to apparent mobilization}

The strike analyses connect aggregate framing to observed contributors. Returning Reddit authors more often mention protest, but their changed expression is only one component of the aggregate increase. Entrants add protest-language records, exits remove earlier contributions, and changes in record weights alter the contribution of returning authors. Neither an aggregate share nor an ever-mentioned indicator isolates changes in a person's beliefs. The activity-restricted cohort further shows why opportunities to observe a label matter: contributing more records makes at least one match more likely.

This extends frame-centred accounts of movement communication without replacing them. \citet{chen2023} compares movement and media framing, and \citet{suitner2023} traces semantic connections during FridaysForFuture. Reddit research examines discussion communities and changes in language \citep{treen2022,fariello2025}. Our decomposition adds an account of which observed participant components produce a frame-share change. It can distinguish expression-rate change from turnover, while leaving the meanings and causes of those contributions unresolved.

Climate Twitter's strike-window collapse illustrates why observation boundaries must precede a social interpretation. The main contributing archive ends before the strike. Its low return rate therefore cannot establish a departure of climate advocates, an event-driven silence or a migration to another platform. We withhold the corresponding inference. \citet{felt2016} and \citet{maemura2023} place collection and extraction practices within the research process; here, those practices determine whether an author trajectory is interpretable at all. Processing every available record improves the accounting of the supplied archive but does not repair its missing event-period coverage.

\subsection{Scale does not resolve dependence or ethical obligations}

The inferential record is narrower than the descriptive record. Most crossed author/month quantities lack released intervals because of limited cluster support or concentrated influence. Joint response covariance estimates fail, so separate author- and month-clustered sensitivities cannot establish valid joint uncertainty. The selected Twitter mitigation/cost association loses adjusted support under calendar-month HAC checks. Exact paired-author tests assume independent discordant accounts, which coordinated participation may violate. These restrictions concern the effective information and dependence structure, not merely the number of archived records. The post-inspection support screen is exploratory and does not guarantee inferential coverage.

Other limitations follow from the archive design. The Twitter collections can overlap, unknown authors are excluded from clustered inference, and observation periods differ by source. Current evidence does not establish representativeness of platform users or of the general public. Narrower phrase rules remain lexical definitions because no independent stance validation was completed. Event windows and fixed-calendar placebos do not identify isolated event effects. Full capture-history verification and a valid joint dependence model would strengthen the existing estimates; endorsement, exposure and offline action would require additional data rather than a more elaborate interpretation of the same counters.

Data reuse also carries obligations separate from inferential quality. \citet{metcalf2016} examines the ethics divide in big-data research, and \citet{gilbert2023} shows that expectations about social media reuse depend on research context and consent. Aggregate reporting and exclusion of raw texts from the manuscript package reduce disclosure but do not establish the legitimacy of the original collection or resolve institutional review. The author must document permissions and an ethics approval or waiver before submission. The study's emphasis on archive provenance must apply to that institutional record as well as to file reconciliation.

\section{Conclusion}

Twitter and Reddit archives can reveal how climate-action and backlash-related words appear together, how those categories relate to recorded responses, and how observed participation changes around events. Those claims require separate denominators, outcomes and observation boundaries. In this study, lexical-definition changes alter source and event contrasts, eligible principal forecasts provide no improvement from cross-discourse history, and participant turnover contributes to protest-language growth. The resulting account is useful because it specifies what the archives establish and where inference must stop. Climate backlash as an attitude, an experienced reaction or a causal consequence of advocacy remains a different empirical question.